\documentclass[times,twocolumn,final]{elsarticle}

\usepackage{prletters}
\usepackage{framed,multirow}

\usepackage{latexsym}
\usepackage{amsmath}
\usepackage{amssymb}

\usepackage{times}
\usepackage{epsfig}
\usepackage{graphicx}

\usepackage{multirow}
\usepackage{adjustbox}

\definecolor{newcolor}{rgb}{.8,.349,.1}

\usepackage[pagebackref=true,breaklinks=true,letterpaper=true,colorlinks,bookmarks=false]{hyperref}

\journal{Pattern Recognition Letters}

\begin{document}

\clearpage
\thispagestyle{empty}
\ifpreprint
  \vspace*{-1pc}
\fi

\clearpage

\ifpreprint
  \setcounter{page}{1}
\else
  \setcounter{page}{1}
\fi

\begin{frontmatter}

\title{Multi-Scale Weight Sharing Network for Image Recognition}

\author[1,2]{Shubhra \snm{Aich}\corref{cor1}}
\cortext[cor1]{Corresponding author:
  E-mail.: s.aich@usask.ca}
\author[1]{Ian \snm{Stavness}}
\author[2]{Yasuhiro \snm{Taniguchi}}
\author[2]{Masaki \snm{Yamazaki}}

\address[1]{Department of Computer Science, University of Saskatchewan, Canada}
\address[2]{Honda R\&D Innovation Lab, Akasaka, Tokyo, Japan}

\received{1 May 2013}
\finalform{10 May 2013}
\accepted{13 May 2013}
\availableonline{15 May 2013}
\communicated{S. Sarkar}

\begin{abstract}
In this paper, we explore the idea of weight sharing over multiple scales in convolutional networks. Inspired by traditional computer vision approaches, we share the weights of convolution kernels over different scales in the same layers of the network. Although multi-scale feature aggregation and sharing inside convolutional networks are common in practice, none of the previous works address the issue of convolutional weight sharing. We evaluate our weight sharing scheme on two heterogeneous image recognition datasets -- ImageNet (object recognition) and Places365-Standard (scene classification). With approximately 25\% fewer parameters, our \emph{shared weight} ResNet model provides similar performance compared to baseline ResNets.
Shared-weight models are further validated via transfer learning experiments on four additional image recognition datasets -- Caltech256 and Stanford 40 Actions (object-centric) and SUN397 and MIT Inddor67 (scene-centric). Experimental results demonstrate significant redundancy in the vanilla implementations of the deeper networks, and also indicate that a shift towards increasing the receptive field per parameter may improve future convolutional network architectures.
\end{abstract}


\end{frontmatter}


\section{Introduction}
In recent years, there has been a steady increase in the depth of deep learning architectures to perform image recognition on massive and more challenging datasets. From $\sim$ 10-20 layers AlexNet \cite{alexnet} and VGG \cite{vgg} models, it is now possible to train models with more than 100 layers, such as ResNet \cite{resnet} and DenseNet \cite{densenet}. In addition, the computation of convolutional features has increased in complexity from straightforward sequential patterns to shortcut connections \cite{resnet} and multi-scale feature aggregation \cite{inception-v1,inception-v3}. Although multi-scale features should be well captured by increasing the depth of convolutional networks, explicit methods have been employed to extract features at multiple scales, including convolution with variable sized kernels \cite{inception-v1} and atrous convolution \cite{deeplab-v1-2,deeplab-v3}. In these methods, multi-scale feature maps are most commonly concatenated along their depth or aggregated element-wise. These explicit multi-scale methods have been shown to work well in practice. All such previous approaches, however, have used separate kernel weights for different scales, rather than sharing weights across scales.

Prior to deep learning, one of the most influential feature extraction algorithms was the Scale-Invariant Feature Transform (SIFT)~\cite{sift,vlfeat}.
Scale-space representations \cite{scale-space-lindeberg} and oriented gradient histograms are the core ideas underlying SIFT. According to scale-space theory, to accommodate the need for feature detection at multiple scales, in SIFT, a difference of Gaussians (DoG) at multiple scales are used to find the local extrema over the input image. These points are then taken as the candidate keypoints for further refinement and descriptors (or oriented gradient histograms) are computed for different applications, such as keypoint matching, classification with bag-of-words, and multi-kernel support vector machines \cite{visual-vocabulary}. The same set of analytical expressions or algorithms (i.e., extrema detection and gradient histogram computation) are used throughout different scales of the input image. Using the same formulae for transformation over different scales of the input provides a better scale-invariant approximation of the desired mapping. This is important because the correct scale of the objects in the image is not known in advance. For this reason, applying the same operators across scales is a central tenant of SIFT and other multi-scale feature descriptors. Within the context of deep learning, we propose that similar properties can be accomplished by sharing weights across scales.

\begin{figure}[]
\centering
\includegraphics[width=0.25\textwidth]{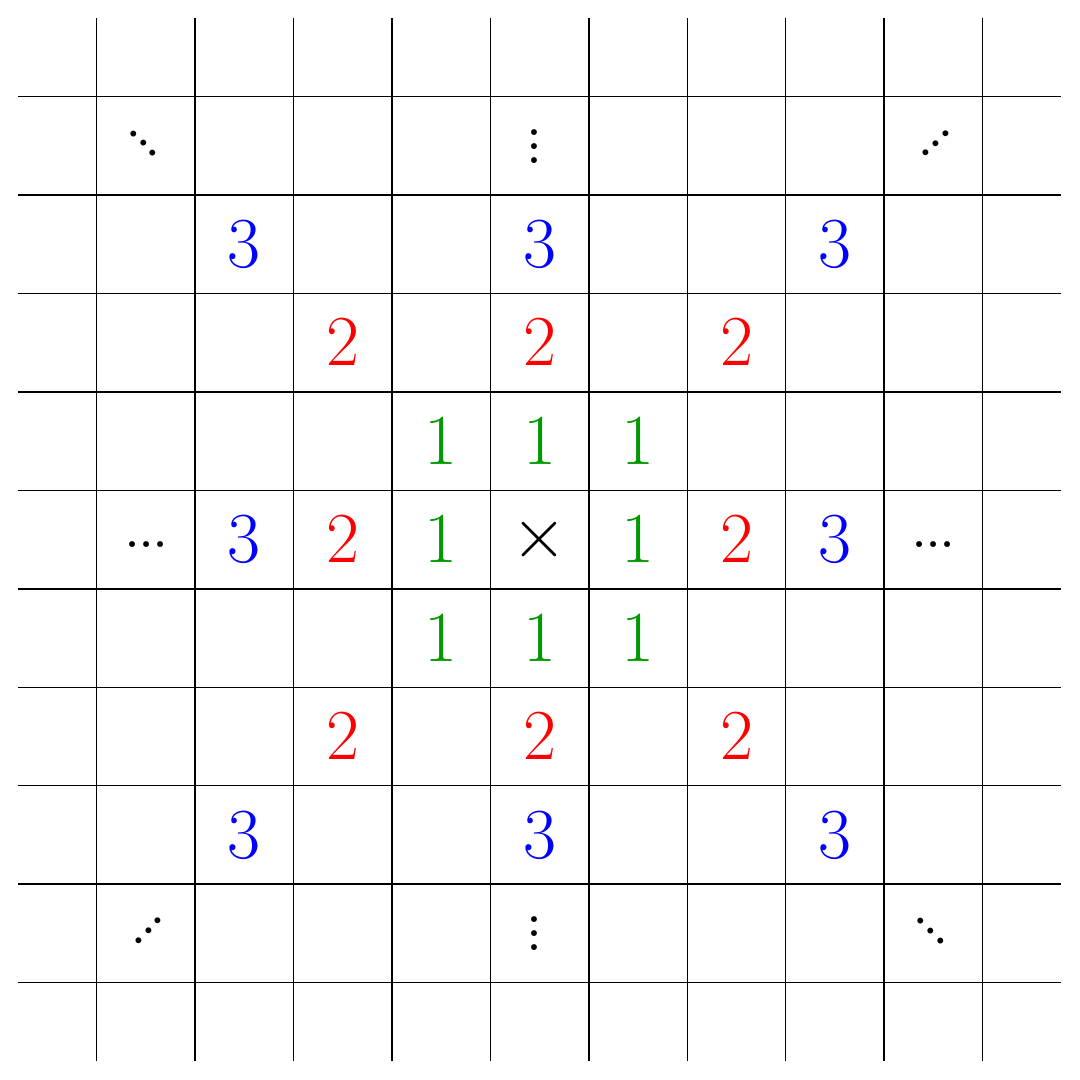}
\caption{$3\times 3$ atrous convolution with multiple rates/scales. Numbers indicate the rate of the convolution. To implement our multi-scale weight sharing scheme, we share the kernels of multiple rates in the same layer of our architecture.}
\label{fig:atrous-conv}
\end{figure}

With the emergence of large-scale datasets and gradient-based optimization frameworks, like deep learning, almost all the scales that might be present in practice can now be covered and learned.
Nonetheless, using the same set of weights over different scales is desirable, both conceptually and pragmatically. By using the same kernel weights, we obtain similar feature map responses regardless of the scale of the input. This causes the model to learn kernel weights in a more scale-invariant manner. We expect that weight sharing amongst kernels of multiple scales should effectively realize the \emph{scale-invariant feature transform} within a deep network architecture.

The motivation for this paper is to employ the idea of weight sharing among the convolution kernels of the same layer of a network based on the inspiration from SIFT (and scale-space representation in general). We hypothesize that a learnable implementation of the scale-space mechanism with convolutional kernels would assist the convolutional architectures in deriving better scale-invariant features. In order to share kernel weights for multiple scales, the shared tuple of kernels are required to have the same shape and different kernels in the shared tuple should be applied to different scales. Atrous convolution \cite{deeplab-v1-2} with variable rates satisfies these constraints (Figure \ref{fig:atrous-conv}). Hence, we share the weights of atrous convolution kernels in the same layer of computation.


We evaluate our weight-sharing scheme on two large-scale, heterogeneous image recognition datasets -- ImageNet \cite{imagenet} and Places365-Standard \cite{places}. On both these datasets, with $\sim 25\%$ fewer parameters and the same computational complexity, weight sharing provides marginally better performance than the corresponding baseline ResNet \cite{resnet} architectures. We also benchmark the effectiveness of the shared-weight features against baselines for transfer learning on four standard object-centric and scene-centric datasets, namely Caltech256 \cite{caltech256}, Stanford 40 Actions \cite{actions40}, SUN397 \cite{sun397}, and MIT Indoor67 \cite{indoor67}. On all these datasets, the shared variant of larger ResNets perform just as well as their baseline counterparts, despite the substantial reduction in parameters. These experimental results both validate our idea of weight sharing and also indicate the redundancy of parameters in existing baseline architectures. 

\section{Related Works}
Our work can be categorized as an architectural enhancement for convolutional networks, therefore we will provide a brief account of related architectural enhancements over the last few years. Next, we will compare our principled approach for parameter reduction through weight sharing to related works that realize parameter reduction through network pruning.

\textbf{Progression of architectural enhancements: }LeNet \cite{lenet} was the first successful example of a convolutional network architecture that was trained with gradient-descent. In this network, the authors demonstrated the effectiveness of weight sharing via sparse convolutional connections rather than fully connected (FC) form. However, it took until 2012 for computer vision researchers to successfully train a deep network for large-scale image classification problems, due to the design constraints (vanishing gradients with sigmoid nonlinearity) and the lack of computational resources. In 2012, AlexNet \cite{alexnet} made a breakthrough by successfully training a moderately large stack of convolutional layers with two very large FC layers on ImageNet \cite{imagenet}. However, it is limited in its convolutional feature extraction capability due to various kernel sizes at different layers of the model. This is later addressed in VGG \cite{vgg} paper, with only $3 \times 3$ convolutions. This design principle makes the architecture easy to adapt and it is still widely used in practice. Also, due to the emergence of effective weight initialization schemes based on the fan-in/fan-out ratio \cite{xavier,he}, both AlexNet and VGG are now amenable to be trained from scratch without any initial warm-up.

The NIN \cite{nin} paper replaces $n \times n$ linear convolution with a multi-layer perceptron in search for a more powerful non-linear approximator. From another perspective, the NIN paper demonstrates a paradigm shift from the heavy-tail architecture to the heavy-head ones replacing large FC layers with the global average pooling (GAP) operation at the backend of the network. This substitution pushes the convolutional front-end of the latter architectures to develop stronger representations alongside the increase in network depth with less overfitting.

According to the standard model of the primate visual cortex \cite{theory-object-detection,cortex-poggio}, the ventral stream first achieves scale and position invariance followed by other transformations with an increase in receptive field with the depth of processing hierarchy. Also, rapid presentation or immediate recognition is hypothesized to be a feedforward task with no scope for further attention. In this regard, both AlexNet and VGG are instantiations of the immediate or sudden recognition pipeline under the theme of convolutional networks. Although ideally, increased depth should cover the full aspect of multi-scale input processing, a better network engineering should boost the performance of CNN, which was addressed in the Inception family of architectures \cite{inception-v1,inception-v3,xception}. The vanilla Inception module \cite{inception-v1} simply concatenates feature representations from variable sized kernels. Later versions include separable convolution \cite{inception-v3} and depthwise separable convolution \cite{xception} for computational efficiency. Next, ResNet \cite{resnet} joins the league of architectures using residual connections to facilitate learning with comparatively much larger depth.

\begin{table}[t!]
\centering
\caption{Architectural comparison of our multi-scale shared ResNet101  with vanilla ResNet101}
\label{tab:architecture}
\begin{adjustbox}{max width=0.5\textwidth,center}
\begin{tabular}{c|c|c|c}
\hline
\multicolumn{1}{c|}{Output} & \multicolumn{1}{c|}{ResNet101} & \multicolumn{1}{c|}{ResNet101(shared)} & \#Blocks \\ \hline
112 & $7\times 7$, 64, /2 & $7\times 7$, 64, /2 & 1 \\ \hline
56 & $3\times 3$ max-pool, /2 & $3\times 3$ max-pool, /2 & 1 \\ \hline
56 & \begin{tabular}[c]{@{}c@{}}$1\times 1$, 64\\ $3\times 3$, 64\\ $1\times 1$, 256\end{tabular} & \begin{tabular}[c]{@{}c@{}}$1\times 1$, 64\\ $3\times 3$, 32, \; $rate \in [1,2],\; shared$ \\ $1\times 1$, 256\end{tabular} & 3 \\ \hline
28 & \begin{tabular}[c]{@{}c@{}}$1\times 1$, 128\\ $3\times 3$, 128\\ $1\times 1$, 512\end{tabular} & \begin{tabular}[c]{@{}c@{}}$1\times 1$, 128\\ $3\times 3$, 64, \; $rate \in [1,2],\; shared$ \\ $1\times 1$, 512\end{tabular} & 4 \\ \hline
14 & \begin{tabular}[c]{@{}c@{}}$1\times 1$, 256\\ $3\times 3$, 256\\ $1\times 1$, 1024\end{tabular} & \begin{tabular}[c]{@{}c@{}}$1\times 1$, 256\\ $3\times 3$, 128, \; $rate \in [1,2],\; shared$ \\ $1\times 1$, 1024\end{tabular} & 23 \\ \hline
7 & \begin{tabular}[c]{@{}c@{}}$1\times 1$, 512\\ $3\times 3$, 512\\ $1\times 1$, 2048\end{tabular} & \begin{tabular}[c]{@{}c@{}}$1\times 1$, 512\\ $3\times 3$, 256, \; $rate \in [1,2],\; shared$ \\ $1\times 1$, 2048\end{tabular} & 3 \\ \hline
\#Parameters & $42.50M$ & $31.83M$ &  \\ \hline
\end{tabular}
\end{adjustbox}
\end{table}

Apart from these recognition architectures, some other computational blocks emerged from the sub-domain of semantic segmentation, object detection, instance segmentation, and object counting. These computational blocks mostly focus on convolutional feature sharing, and include atrous or hole convolution \cite{deeplab-v1-2,deeplab-v3,deeplab-v3+}, spatial pyramid pooling \cite{spp}, ROI-pooling \cite{fast-rcnn,faster-rcnn}, ROI-align \cite{mask-rcnn}, global sum pooling \cite{gsp}, etc.
However, although convolutional \emph{feature sharing} in different forms is prevalent in the deep learning community, no work has been published yet exploring the idea of convolutional \emph{weight sharing} in any form.
\\ ~ \\
\textbf{Weight-sharing \textit{vs.} pruning: }We differentiate our principled approach of weight-sharing from pruning-based approaches for reducing network parameters in a number of ways. First, in this paper, we try to emulate the SIFT-like multi-scale computation with negligible computational overhead, where unlike pruning, parameter reduction is an added by-product. Second, from the Optimal Brain Damage paper \cite{optimal-brain-damage} to recent neural pruning \cite{neural-pruning} methods, pruning is done either on the trained models or selectively on part of the available weights, whereas we train fewer (and the same set of) parameters with a lower risk of overfitting from the very beginning. Moreover, pruning on large datasets is computationally prohibitive, which is why recent pruning methods are validated on much smaller CIFAR datasets compared to our experiments with very large models (ResNet101/152) on some of the largest image-recognition datasets to date.

\section{Our Approach}

To share weights between multiple scales, all the scales must have the same number of weights in each layer. One option is to use the same set of kernels for multiple consecutive convolutional blocks. But this would not mimic the idea of multi-scale feature computation with the same weights, because each consecutive layer transforms the feature into a new nonlinear space. Conventional multi-scale feature descriptors extract features from the same input, scaled by multiple factors, thus representing the same input space. Also, we found poor convergence with such a scheme in practice.

Instead, we use the idea of atrous convolution to realize our weight sharing scheme. For each convolutional layer of a deep model, if there are $k_o$ kernels of dimension $k_i \times f \times f$ where $k_i$ is the dimension of the incoming feature maps, and $f$ is the kernel size, we employ each set of $k_o/n$ kernels for convolution with rates $[1,n]$. All of these kernel sets of dimension $k_o/n$ share the same parameters. Thus, the number of parameters in the convolutional layer of the shared models goes down to $k_o/n$ (or, a reduction by $k_o(n-1)/n$). Note that a vanilla deep model is a special case of our multi-scale weight sharing scheme with $n=1$. In initial experiments, we attempted to train models with $n\in \{2,4\}$ and found poor convergence for $n=4$. We believe that the reason for poor performance of models with the range of higher rates is the lack of a sufficient number of unique parameters for those models. Each time we divide the set of $k_o$ kernels into $n$ subsets, the number of model parameters reduces by $k_o/n$. That causes about a $75\%$ reduction in the number of convolutional parameters in the models for $n=4$. Based on these pilot empirical results, we instead choose $n=2$ for our model.

One potential concern about atrous convolution might be that its output should be noisier than the features extracted from image pyramids. This is because atrous convolution computes features from the subsampled input directly without smoothing, whereas gaussian smoothing is used before Laplacian extrema detection in SIFT. However, the advantage of atrous convolution over LoG/DoG filters is that the weights of the convolution kernels are learned. Thus, it is possible to learn the kernel parameters that are a convolved version of a smoothing filter like Gaussian and feature encoding filter like edge detection. Moreover, it is a reasonable assumption that such noise would be suppressed in the deeper layers the network, considering the layer-wise progression of features. In this regard, atrous convolution is the best available option to devise a multi-scale parameterization inside CNN architectures.

Weight sharing must be injected carefully into the vanilla models because if it is used in a smaller model like ResNet34 or ResNet50, the number of trainable parameters or the dimensionality of the model is not sufficiently large to capture the data distribution for the task at hand. Consequently, weight sharing should be applied to deeper models, like ResNet101 or ResNet152, where we would have sufficient parameters after reduction. Although a wider network may be used with the number of new kernels $k_o^\prime = nk_o$ for the range of $n-$scale atrous convolution, such modification will linearly increase the computation cost by $\mathcal{O}(n)$. 

In the experiment section, we show that for $n=2$, with approximately $25\%$ fewer trainable parameters, our model shows similar or slightly better performance than the corresponding vanilla model without any atrous convolution, and also the model with atrous convolution but without any weight sharing.

Finally, our weight sharing scheme for a single convolutional layer can be expressed as follows:

\begin{equation}
  \begin{aligned}
& f_o^k(t) = f_i(t) * \omega (t) \vert_{r=k}; \;\;\; k=1,2,\hdots,n \\
& \varDelta^k(t+1) = \frac{\partial \mathcal{L}(t) }
{\partial f_o^k(t)}
\frac{\partial f_o^k(t)} {\partial \omega(t)} \\
& \varDelta(t+1) = \mathcal{E}[\{\varDelta^k(t+1)\}] \\
& \omega(t+1) = \omega(t) - \eta \varDelta(t+1) \\
  \end{aligned}
\end{equation}

Here, $f_i$ and $f_o^k$ denote the input and output feature maps, respectively, and the superscript indicates the rate of dilation. The expected gradient over the set $\{\varDelta^k(t+1)\}$ is denoted by $\varDelta(t+1)$ that is used for gradient descent.

\section{Experiments}

We use two complementary large-scale image classification datasets to evaluate our idea of multi-scale weight sharing. One of them is for object recognition (ImageNet \cite{imagenet}) and the other is for scene recognition (Places365-Standard \cite{places}). These two datasets are complementary, because for object recognition, the model should focus on a compact region for detail, whereas for scene recognition, capturing the underlying connectivity structure of the scene components (both objects and stuff) is important.

\subsection{Datasets}

\noindent\textbf{ImageNet:} ImageNet \cite{imagenet} is the standard benchmark for new ideas regarding convolutional networks. It contains $1.28M$ images from 1000 object classes and $50K$ validation images with $50$ samples per category.

\noindent \textbf{Places365-Standard:} This is a subset of the original database \cite{places} comprising 10$M$ images. It has 1.8$M$ training images with the images per class in the range $[3068, 5000]$. The validation set contains 100 images per category. We use the small resolution version of the dataset where each image is resized to $256 \times 256$ regardless of the true aspect ratio.

\begin{table}[]
\centering
\caption{Results on ImageNet and Places365-Standard validation sets. To be consistent with previous benchmarks, ImageNet and Places statistics are provided on $224\times 224$, single-crop images, respectively}
\label{tab:accuracy}
\begin{adjustbox}{max width=0.5\textwidth,center}
\begin{tabular}{cl|c|llll}
\hline
\multicolumn{2}{c|}{\multirow{3}{*}{Model}} & \multicolumn{1}{c|}{\multirow{3}{*}{\begin{tabular}[c]{@{}c@{}}\#Params\\ $(\times 10^6)$ \end{tabular}}} & \multicolumn{4}{c}{Error (\%)} \\ \cline{4-7}
\multicolumn{2}{c|}{} & \multicolumn{1}{c|}{} & \multicolumn{2}{c|}{ImageNet} & \multicolumn{2}{c}{Places365} \\ \cline{4-7}
\multicolumn{2}{c|}{} & \multicolumn{1}{c|}{} & \multicolumn{1}{c|}{Top-1} & \multicolumn{1}{l|}{Top-5} & \multicolumn{1}{l|}{Top-1} & Top-5 \\ \hline
\multirow{3}{*}{AlexNet} & Vanilla & 57.00 & \multicolumn{1}{l|}{40.86} & \multicolumn{1}{l|}{18.74} & \multicolumn{1}{l|}{48.29} & 17.98 \\
 & Unshared & 57.00 & \multicolumn{1}{l|}{41.51} & \multicolumn{1}{l|}{19.32} & \multicolumn{1}{l|}{48.57} & 18.30 \\
 & Shared & 55.77 & \multicolumn{1}{l|}{44.36} & \multicolumn{1}{l|}{21.52} & \multicolumn{1}{l|}{49.63} & 19.32 \\ \hline
\multirow{3}{*}{ResNet18} & Vanilla & 11.17 & \multicolumn{1}{l|}{30.24} & \multicolumn{1}{l|}{10.92} & \multicolumn{1}{l|}{45.87} & 15.28 \\
 & Unshared & 11.17 & \multicolumn{1}{l|}{29.21} & \multicolumn{1}{l|}{10.45} & \multicolumn{1}{l|}{45.53} & 15.39 \\
 & Shared & 8.04 & \multicolumn{1}{l|}{31.43} & \multicolumn{1}{l|}{11.45} & \multicolumn{1}{l|}{45.88} & 15.68 \\ \hline
\multirow{3}{*}{ResNet34} & Vanilla & 21.29 & \multicolumn{1}{l|}{25.97} & \multicolumn{1}{l|}{8.34} & \multicolumn{1}{l|}{44.71} & 14.72 \\
 & Unshared & 21.29 & \multicolumn{1}{l|}{26.22} & \multicolumn{1}{l|}{8.43} & \multicolumn{1}{l|}{44.51} & 14.52 \\
 & Shared & 15.63 & \multicolumn{1}{l|}{27.33} & \multicolumn{1}{l|}{9.04} & \multicolumn{1}{l|}{44.93} & 14.91 \\ \hline
\multirow{3}{*}{ResNet50} & Vanilla & 23.51 & \multicolumn{1}{l|}{23.68} & \multicolumn{1}{l|}{6.90} & \multicolumn{1}{l|}{44.06} & 14.16 \\
 & Unshared & 23.51 & \multicolumn{1}{l|}{23.17} & \multicolumn{1}{l|}{6.68} & \multicolumn{1}{l|}{43.96} & 13.71 \\
 & Shared & 17.85 & \multicolumn{1}{l|}{23.55} & \multicolumn{1}{l|}{6.97} & \multicolumn{1}{l|}{44.26} & 14.08 \\ \hline
\multirow{3}{*}{ResNet101} & Vanilla & 42.50 & \multicolumn{1}{l|}{22.63} & \multicolumn{1}{l|}{6.44} & \multicolumn{1}{l|}{43.68} & 13.70 \\
 & Unshared & 42.50 & \multicolumn{1}{l|}{21.90} & \multicolumn{1}{l|}{6.03} & \multicolumn{1}{l|}{43.59} & 13.50 \\
 & Shared & 31.83 & \multicolumn{1}{l|}{22.07} & \multicolumn{1}{l|}{6.31} & \multicolumn{1}{l|}{43.65} & 13.67 \\ \hline
\multirow{3}{*}{ResNet152} & Vanilla & 58.14 & \multicolumn{1}{l|}{21.64} & \multicolumn{1}{l|}{5.77} & \multicolumn{1}{l|}{43.99} & 13.98 \\
 & Unshared & 58.14 & \multicolumn{1}{l|}{21.63} & \multicolumn{1}{l|}{5.89} & \multicolumn{1}{l|}{43.73} & 13.61 \\
 & Shared & 43.34 & \multicolumn{1}{l|}{22.07} & \multicolumn{1}{l|}{6.25} & \multicolumn{1}{l|}{43.86} & 13.83 \\ \hline
\end{tabular}
\end{adjustbox}
\end{table}

\subsection{Implementation}
We implement our weight sharing scheme in PyTorch \cite{pytorch}. As the architecture, we modify vanilla ResNet models (Table \ref{tab:architecture}) and AlexNet from the old-fashioned networks. Initial experimentation was performed with multiple rates, but we found better accuracy with $rate \in [1,2]$. As already mentioned, for higher rates, we attribute the poor performance to the lack of sufficient number of parameters. Although the number of parameters can be kept constant by increasing the depth, we did not do this because it would increase the computation by a non-negligible extent compared to the vanilla implementation. Our goal is to reduce the number of trainable parameters without increasing the computational burden compared to the existing architectures.

\subsection{Training}
The models are trained on 2 NVIDIA workstations -- one comprising 4 TITAN Xp GPUs and the other containing 4 Tesla K80s. To be consistent with the previous benchmarks, we follow the same hyperparameter settings. We use SGD as the optimizer with initial learning rate, momentum, and weight decay of 0.1, 0.9, and 0.0001, respectively. All of the models are trained for 100 epochs with a batch size of 256, except for the ResNet152 variants where we used a batch size of 192 to accommodate the higher memory requirement. The learning rate is multiplied by 0.1 after every 30 epochs.

\begin{figure}[h!]
\centering
\includegraphics[width=0.4\textwidth]{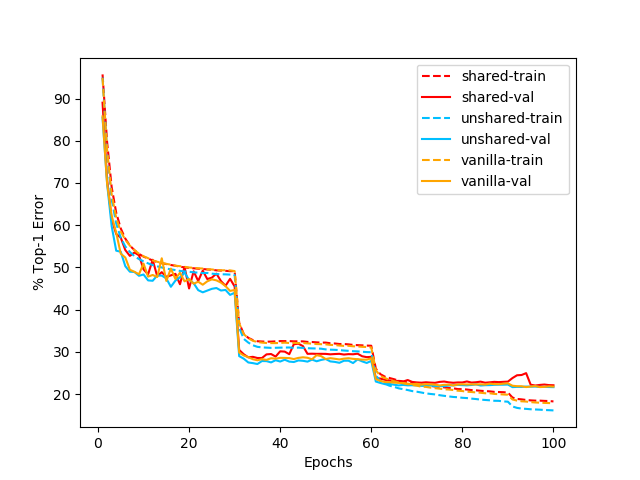} \\
\includegraphics[width=0.4\textwidth]{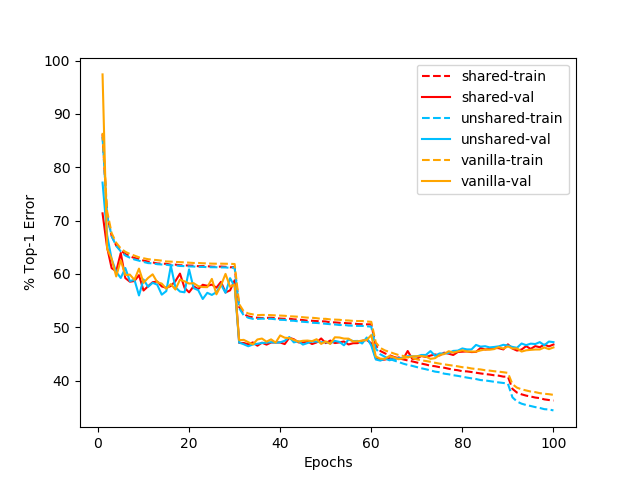}
\caption{Training curve comparisons for ResNet152 architectures: ImageNet(top) and Places(bottom). All the variants exhibit very similar training statistics. Best viewed in digital format.}
\label{fig:train-curve}
\end{figure}

\subsection{Results}
In Table \ref{tab:accuracy}, we present the results for both ImageNet and Places365-Standard datasets with the number of parameters in each model. The number of parameters in the final classification layer is not taken into account because it is small compared to the rest of the model and also does not contribute to the comparison, in general. Moreover, single crop ($224 \times 224$) error rates are reported here following the previous literature.

In this table, AlexNet-shared performance is significantly lower than AlexNet-unshared, which is also inferior than vanilla AlexNet. Because AlexNet has only few convolutional layers, reducing the number of parameters in those layers results in a poor feature extractor with insufficient capacity. For smaller ResNets (ResNet18, ResNet34, and ResNet50), shared versions almost always exhibit slightly inferior performance than the vanilla counterparts.
For these smaller variants of ResNet, we hypothesize that the parameter reduction in smaller models causes an influential reduction in the overall capacity of the model in the function space, which is measured with the VC dimension \cite{vapnik-stat-learning-theory} in statistical learning theory.

For larger models (ResNet101 and ResNet152), unshared variants are marginally better than both vanilla and shared, and the shared ones demonstrate slight gain over the vanilla implementation. Therefore, for larger models, we argue that the model with shared weights is at least comparable to the existing alternatives in terms of performance, which is also evident from the training and validation statistics shown in Figure \ref{fig:train-curve}.

There are a few more complicated variants of the vanilla ResNet architecture, such as ResNeXt \cite{resnext} and Wide-ResNet \cite{wide-resnet}. All these specializations convey the same principle as their parent architecture with significantly high similarity to the vanilla ResNets. For this reason, we expect that the weight sharing would have a similar effect on all these updated, more complex realizations, and so, we have chosen to study the effect of weight sharing only on the standard ResNet of different depths, in detail.

By sharing weights, we reduce the number of parameters in the model by about $25\%$ without reducing the model's accuracy (less than $0.2\%$ difference in accuracy for unshared vs. shared). In other words, for an increase in the number of trainable parameters by $\sim 25\%$, we get very little or almost no improvement. This manifests the redundancy or saturation in the architectural design of the original implementation. According to the principle of least privilege as well as minimum description length (MDL), we hypothesize that removing such additional parameters might give better generalization performance in a more critical scenario. Also, this results suggests that we do not need more parameters than those of the existing architectures to improve performance. Rather, we need to increase the computation or receptive field per parameter.

Note that, we started training DenseNet161 \cite{densenet} for the Places dataset, but found that it had a much slower convergence rate as compared to the ResNet models during the first few epochs. Hence, we did not continue training further. The slower convergence of DenseNet is possibly due to the lack of sufficient growth rate necessary for the more complicated \textit{Places} dataset.

Moreover, in \cite{places}, ResNet152 was finetuned on the \textit{Places} dataset only. However, we train this model also from scratch. Comparing its performance with ResNet101 in Table \ref{tab:accuracy}, we hypothesize that ResNet152 ($\sim 58M$ parameters) might suffer from a little bit of overfitting on the \textit{Places} dataset.

\begin{figure}[t!]
\centering
\includegraphics[width=0.23\textwidth]{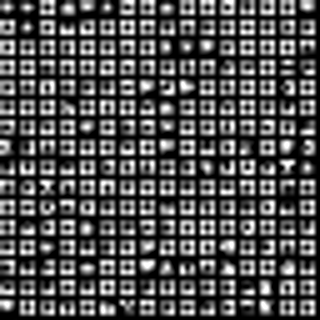}
\vspace{0.5em}
\includegraphics[width=0.23\textwidth]{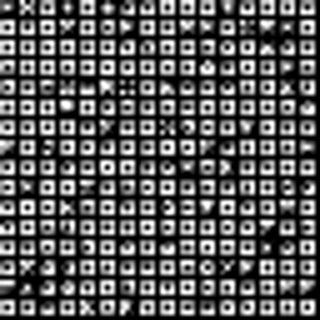}
\caption{Highest activated 256 $3\times 3$ kernels based on the aggregate magnitude of the weight parameters in the last convolutional layer of vanilla ResNet101-vanilla (left) and ResNet101-shared (right) trained on ImageNet dataset. Both models exhibit very similar patterns of weights.}
\label{fig:weights-imagenet}
\end{figure}

\begin{figure}[t!]
\centering
\includegraphics[width=0.23\textwidth]{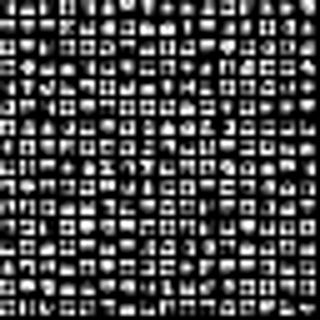}
\vspace{0.5em}
\includegraphics[width=0.23\textwidth]{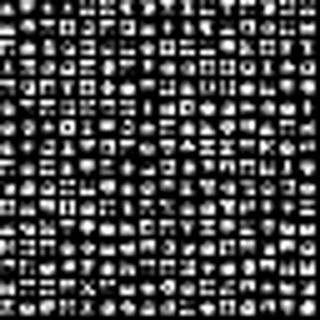}
\caption{Highest activated 256 $3\times 3$ kernels based on the aggregate magnitude of the weight parameters in the last convolutional layer of vanilla ResNet101-vanilla (left) and ResNet101-shared (right) trained on Places365-Standard dataset. Like ImageNet models, both models also develop similar weight patterns on this dataset, but different from those on ImageNet.}
\label{fig:weights-places}
\end{figure}

For exploratory visualization, we plot the top 256 kernels in the last convolutional layer of the vanilla and shared ResNet101 models based on the higher aggregate magnitude of the weight parameters for both ImageNet (Figure \ref{fig:weights-imagenet}) and \textit{Places} (Figure \ref{fig:weights-places}). From these figures, we observe a similar pattern of kernel weights for both models on each dataset. Cross-datasets patterns are different which is expected because of their heterogeneous nature. The similarity of intra-dataset kernel patterns is another illustration of the parameter redundancy in the vanilla ResNet101 architecture, which can be minimized by our proposed weight sharing scheme.

Another observation already addressed in \cite{places} is the somewhat comparable accuracy of the old-fashioned architectures like AlexNet or VGG with the recent ones on the \textit{Places} database. The accuracy of these models on ImageNet is low compared to ResNets and DenseNets partly due to overfitting because of the large number of parameters. Also, from the perspective of feature encoding, the comparatively lower depth (number of convolutional layers) in VGG and AlexNet is not sufficient for the necessary encoding of detailed textures, which is particularly true for object-centric datasets like ImageNet. However, for scene recognition, detailed information about any single object instance in the image is not necessary. Instead, only a rough or gross impression of many distinguishing or characteristic objects and stuff from the whole image is needed. To obtain that level of information, the number of convolutional layers in AlexNet and VGG is sufficient.

Another aspect of the difference in performance of AlexNet on these two datasets is the probable effect of FC layers. FC layers amalgamate information from the whole input region. For object-centric datasets like ImageNet, the object of interest is mostly located around the central region of the image. However, in a single image, there might be other prominent object candidates elsewhere in the images. Now, when the convolutional feature maps are mixed by the FC layers, features from all of these object candidates are aggregated and that leads to a poorer top-1 error for object recognition. Like the object-centric images in ImageNet, the scene images in the \textit{Places} database are also collected from the web with specific category names. For scene recognition, the best view of that particular scene, comprising an arbitrary number of objects, is taken into account. Therefore, to categorize each image belonging to a particular scene, the model needs to focus on different objects present in the image to get an accurate prediction of the scene category. This means that the convolutional features over the whole image need to be combined well for scene recognition. The FC layers of VGG and AlexNet assist in this information blending and provide comparable performance on the \textit{Places} dataset even without a large number of convolutional layers.

\subsection{Transfer learning}

In the previous section, we demonstrate the effectiveness of weight sharing inside ConvNets on ResNet variants trained from scratch on two large image recognition datasets (ImageNet-1.2M and Places-1.8M). We also expect that weight-sharing may be useful in the context of transfer learning with smaller datasets. Here, we investigate the comparative effectiveness of vanilla, unshared, and shared features for transfer learning on four additional datasets -- two for scene recognition (\textit{SUN397} and \textit{MIT-Indoor67}) and two for object recognition (\textit{Caltech256} and \textit{Actions40}).

\subsubsection{Datasets for transfer learning}

\noindent\textbf{SUN397 \cite{sun397}:}  Scene UNderstanding (SUN) was the largest scene database before \textit{Places}. It contains 397 categories, collected mostly from WordNet \cite{wordnet}, except a few additional scenes, such as \textit{jewelry store} and \textit{mission}. Images were collected via search engines. All the samples are at least of $200\times 200$ resolution. Duplicate, grayscale, blurry, distorted, and noisy images are removed carefully. Also, images for similar scene categories, such as \textit{abbey}, \textit{church}, and \textit{cathedral}, are classified with explicit rules. The final dataset contains $108754$ samples over 397 classes with at least 100 samples for each of the categories. $50-50$ split is done over each class for our experiments.
\\ ~ \\
\noindent\textbf{MIT-Indoor67 \cite{indoor67}:} This dataset comprises indoor scene categories only. Images were collected via an online search engines, Flickr and LabelMe \cite{labelme} dataset. Successful classification of indoor scenes require both global spatial features (e.g. \textit{corridors}) as well as local object-level information (e.g. \textit{bookstores}). The dataset contains 15620 images and 67 categories. However, we have used the train-test split $(67*80/67*20)$ specified in the dataset website for our experiments. Therefore, our training and test set comprise 5360 and 1340 samples, respectively.
\\ ~ \\
\noindent\textbf{Caltech256 \cite{caltech256}:} This object-centric dataset is an enhanced version of Caltech101 \cite{caltech101}, where the enhancements include more categories (256 instead of 101), more images per category (minimum 80), careful avoidance of rotational artifacts, and an additional category for background rejection. Also, unlike Caltech101, no left-right object alignment was performed which makes the new one more natural and challenging. Images were acquired from Google and PicSearch, and duplicates were removed by matching SIFT descriptors. The final dataset consists of 30607 images of which $50-50$ random split is done over each category for our experiments similar to the SUN397 dataset.
\\ ~ \\
\noindent\textbf{Actions40 \cite{actions40}:} The Stanford 40 Actions dataset is another object-centric (or human-centric) dataset comprising 40 diverse actions as the categories, such as \textit{cooking}, \textit{fixing a car}, \textit{walking the dog}, etc. Similar to other datasets, the images were also collected from Google, Bing, and Flickr. For our experiment, we used predefined train-test split, and cropped the images using the bounding box annotation, which is provided for the person performing the target action to avoid confusion. Training and test subsets consists of 4000 and 5532 samples, respectively.

\begin{table}[]
\centering
\caption{Transfer learning results on object-centric datasets with ImageNet pretrained models}
\label{tab:object-ft}
\begin{adjustbox}{max width=0.49\textwidth,center}
\begin{tabular}{cl|cccc}
\hline
\multicolumn{2}{c|}{\multirow{3}{*}{Model}} & \multicolumn{4}{c}{Error (\%)} \\ \cline{3-6}
\multicolumn{2}{c|}{} & \multicolumn{2}{c|}{Caltech256} & \multicolumn{2}{c}{Actions40} \\ \cline{3-6}
\multicolumn{2}{c|}{} & \multicolumn{1}{c|}{Top-1} & \multicolumn{1}{l|}{Top-5} & \multicolumn{1}{l|}{Top-1} & Top-5 \\ \hline
\multirow{3}{*}{ResNet101} & Vanilla & \multicolumn{1}{c|}{13.18} & \multicolumn{1}{c|}{3.82} & \multicolumn{1}{l|}{36.28} & 11.10 \\
 & Unshared & \multicolumn{1}{c|}{13.28} & \multicolumn{1}{c|}{3.56} & \multicolumn{1}{l|}{35.20} & 10.59 \\
 & Shared & \multicolumn{1}{c|}{13.25} & \multicolumn{1}{c|}{3.78} & \multicolumn{1}{l|}{35.77} & 11.59 \\ \hline
\multirow{3}{*}{ResNet152} & Vanilla & \multicolumn{1}{c|}{12.38} & \multicolumn{1}{c|}{3.39} & \multicolumn{1}{l|}{34.24} & 10.41 \\
 & Unshared & \multicolumn{1}{c|}{12.64} & \multicolumn{1}{c|}{3.48} & \multicolumn{1}{l|}{34.27} & 10.38 \\
 & Shared & \multicolumn{1}{c|}{12.70} & \multicolumn{1}{c|}{3.70} & \multicolumn{1}{l|}{35.14} & 10.59 \\ \hline
\end{tabular}
\end{adjustbox}
\end{table}

\begin{table}[]
\centering
\caption{Transfer learning results on scene-centric datasets with \textit{Places} pretrained models}
\label{tab:scene-ft}
\begin{adjustbox}{max width=0.49\textwidth,center}
\begin{tabular}{cl|cccc}
\hline
\multicolumn{2}{c|}{\multirow{3}{*}{Model}} & \multicolumn{4}{c}{Error (\%)} \\ \cline{3-6}
\multicolumn{2}{c|}{} & \multicolumn{2}{c|}{SUN397} & \multicolumn{2}{c}{MIT-Indoor67} \\ \cline{3-6}
\multicolumn{2}{c|}{} & \multicolumn{1}{c|}{Top-1} & \multicolumn{1}{l|}{Top-5} & \multicolumn{1}{l|}{Top-1} & Top-5 \\ \hline
\multirow{3}{*}{ResNet101} & Vanilla & \multicolumn{1}{c|}{24.33} & \multicolumn{1}{c|}{4.38} & \multicolumn{1}{l|}{14.33} & 1.64 \\
 & Unshared & \multicolumn{1}{c|}{23.90} & \multicolumn{1}{c|}{4.22} & \multicolumn{1}{l|}{14.48} & 1.57 \\
 & Shared & \multicolumn{1}{c|}{24.24} & \multicolumn{1}{c|}{4.39} & \multicolumn{1}{l|}{14.69} & 1.87 \\ \hline
\multirow{3}{*}{ResNet152} & Vanilla & \multicolumn{1}{c|}{23.95} & \multicolumn{1}{c|}{4.38} & \multicolumn{1}{c|}{14.10} & 1.57 \\
 & Unshared & \multicolumn{1}{c|}{23.83} & \multicolumn{1}{c|}{4.14} & \multicolumn{1}{l|}{14.85} & 1.64 \\
 & Shared & \multicolumn{1}{c|}{24.06} & \multicolumn{1}{c|}{4.31} & \multicolumn{1}{l|}{14.40} & 1.94 \\ \hline
\end{tabular}
\end{adjustbox}
\end{table}

\subsubsection{Results for transfer learning}

Table \ref{tab:object-ft} and \ref{tab:scene-ft} list the results for finetuning ImageNet and \textit{Places} pretrained models on two object-centric and scene-centric datasets each, respectively. We use only larger models (ResNet101 and ResNet152), for which parameter sharing does not degrade the performance due to lower number of effective parameters. As evident from these tables, there is no clear advantage in performance for any of the model variants. For example, in Table \ref{tab:scene-ft}, for SUN397 dataset, with ResNet101, $unshared > shared > vanilla$ ($>$ indicates marginally better Top-1 performance), and with ResNet152, $unshared > vanilla > shared$. On the other hand, for Caltech256 dataset, with ResNet101, $vanilla > shared > unshared$, and for ResNet152, $vanilla > unshared > shared$. Furthermore, this marginal ranking will differ if $Top-5$ error is considered, which may be a more reasonable metric when multiple objects are present in the image. Therefore, we argue that all the models exhibit similar performance on these finetuning datasets with comparatively fewer training samples. Consistent with the results for training-from-scratch, the shared-weight variants in transfer learning achieve the same performance with $~25\%$ fewer parameters, and therefore can be considered just as effective while being $~25\%$ more efficient than their counterparts.

\section{Conclusion}

In this paper, we investigate weight sharing over multiple scales in convolutional networks. With the shared weight model, which contains approximately 25\% fewer parameters, we achieve similar accuracy compared to the baseline model on two large-scale image recognition datasets. Also, the shared-weight features are found to be equally strong in transfer learning. Our study demonstrates that there is substantial parameter redundancy in existing architectures that can be leveraged to reduce model size while maintaining performance by our proposed weight sharing scheme. Our experiments also indicate that in order to improve performance, we need to increase the computation or receptive field per parameter rather than the number of parameters itself. Following the principle of least privilege and minimum description length, we expect that parameter reduction by weight sharing will eventually be helpful for better generalization in more complex scenarios, and we plan to investigate this in future work.

\section*{Acknowledgments}
This research was undertaken thanks to Honda R\&D Innovation Lab Tokyo (HIL-TK).

\bibliographystyle{plainnat}
\bibliography{egbib,pruning}

\end{document}